\documentclass[letterpaper, 10 pt, conference]{ieeeconf}  

\IEEEoverridecommandlockouts                              

\usepackage{mathtools}
\usepackage{amsmath, amssymb, dsfont}
\usepackage{bm}
\usepackage{graphics}
\usepackage{subcaption}
\usepackage{cite}
\usepackage{ulem}
\usepackage{url}
\usepackage[dvipsnames]{xcolor}
\usepackage{booktabs}
\usepackage{float}
\usepackage{afterpage}
\usepackage{placeins}

\title{\LARGE \bf
Tactile-based force estimation for interaction control with robot fingers
}

\author{Elie Chelly$^{1}$, Andrea Cherubini, Philippe Fraisse, Faïz Ben Amar, and Mahdi Khoramshahi
\thanks{*This work was supported by ANR (PEPR PC2 PI2-IMM Organic Robotics) under grant 22-EXOD-0003}
\thanks{E. Chelly, M. Khoramshahi and F. Ben Amar are with Institut des sytèmes intelligents et de la robotique, Sorbonne University,
        4 Place Jussieu,75005, Paris, France. P. Fraisse is with LIRMM, Montpellier University. A. Cherubini is with Nantes Université, École Centrale Nantes, CNRS, LS2N, UMR 6004, 1, rue de la Noe, 44321 Nantes, France.}
\thanks{$^{1}$ Corresponding author: 
         {\tt\small chelly@isir.upmc.fr}}
}

\begin{document}

\maketitle
\thispagestyle{empty}
\pagestyle{empty}

\begin{abstract}
Fine dexterous manipulation requires reactive control based on rich sensing of manipulator-object interactions. 
Tactile sensing arrays provide rich contact information across the manipulator's surface. 
However their implementation faces two main challenges: accurate force estimation across complex surfaces like robotic hands, and integration of these estimates into reactive control loops.
We present a data-efficient calibration method that enables rapid, full-array force estimation across varying geometries, providing online feedback that accounts for non-linearities and deformation effects. 
Our force estimation model serves as feedback in an online closed-loop control system for interaction force tracking. 
The accuracy of our estimates is independently validated against measurements from a calibrated force-torque sensor.
Using the Allegro Hand equipped with Xela uSkin sensors, we demonstrate precise force application through an admittance control loop running at 100Hz, achieving up to 0.12$\pm$0.08 [N] error margin—results that show promising potential for dexterous manipulation.
\end{abstract}


\section{INTRODUCTION}
Tactile sensing is key to advancing robotic manipulation capabilities through its rich information about robot-object interactions. 
Research has shown that tactile sensors can provide detailed feedback about object properties, including shape, texture, and stiffness~\cite{luo2017robotic}. 
Beyond physical properties, the rich contact information enables perception of dynamic phenomena like slippage and enhances object assessment in applications such as robotic harvesting~\cite{kitouni2023model,mandil2023tactile}. 
This enables robots to perform complex tasks~\cite{Yu2023precise}, particularly in situations where visual feedback alone is insufficient~\cite{ichiwara2022contact}. 
Rich tactile feedback also enhances robotic grasping by enabling continuous adjustment of interaction forces, based on object properties~\cite{funabashi2022multi}.
Through tactile sensing, robots can leverage information on physical properties, dynamic phenomena, and interaction, enhancing their versatility and enabling them to adapt and operate intelligently in challenging environments.

While tactile sensing provides rich and useful contact information, integrating it into control schemes remains challenging due to the high-dimensional nature of the signal.
Another challenge is the diversity of existing tactile sensing modalities which results in the development of sensor specific solutions.
Approaches to integrate tactile sensing in robotic grasping and manipulation have evolved along two complementary paradigms: \textbf{model-based} and \textbf{model-free} approaches.
In \textbf{model-based methods}, physical features extracted from tactile data, such as friction coefficient~\cite{chen2018tactile}, slippage~\cite{VanAnhHo2010}, object pose~\cite{murali2022active}, and material characteristics~\cite{Dai2022}, form essential components of well-defined models that describe object-robot interactions. 
The interpretability of these methods, rooted in physical principles, provides clear insights into the manipulation process~\cite{Kappassov2015}. 
However, capturing complex, nonlinear sensor responses and geometric deformations remains challenging, often requiring sophisticated calibration processes.

\begin{figure}
    \centering
    \includegraphics[width=1\linewidth]{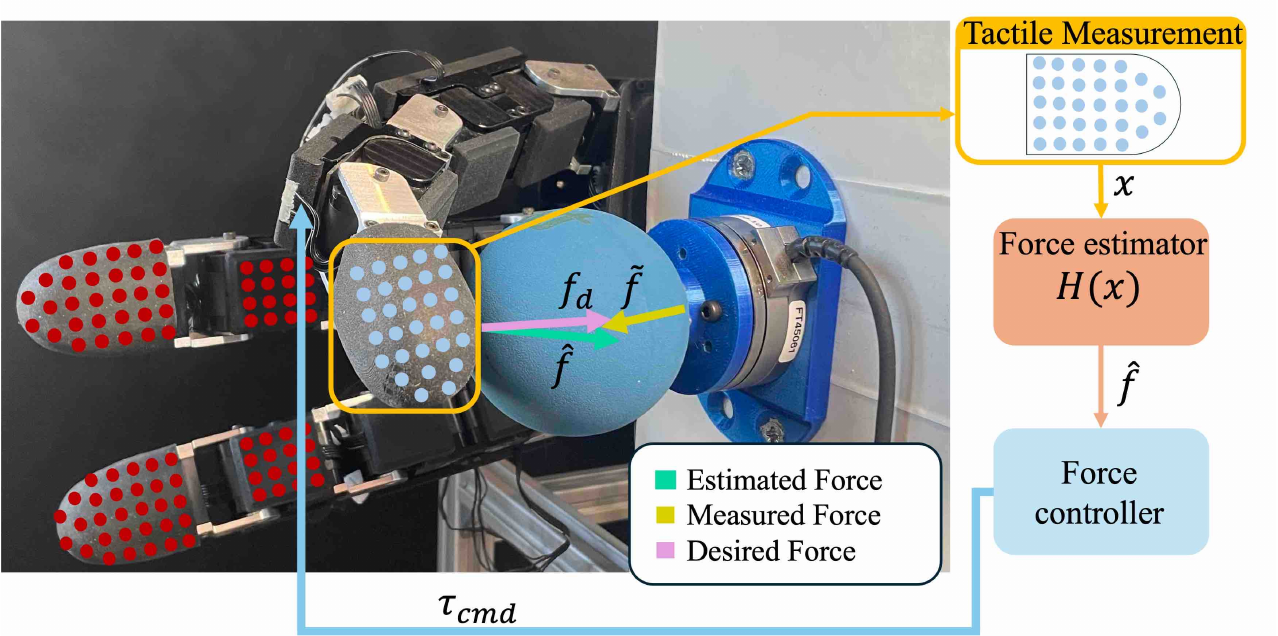}
    \caption{The tactile observation $\boldsymbol{x}$ is mapped by function $\boldsymbol{\mathcal{H}}$ to estimate interaction force $\boldsymbol{\hat{f}}$. This $\boldsymbol{\hat{f}}$ is fed back to the force controller, to adjust the force's orientation and magnitude, ensuring the finger applies desired force $\boldsymbol{f}_{d}$. The measured force $\boldsymbol{\tilde{f}}$ serves as ground truth for the estimator.}
\vspace{-0.7cm}

    \label{fig:intro_fig}
\end{figure}

\textbf{Model-free approaches}, particularly those leveraging deep learning, directly handle high-dimensional tactile data for tasks such as dexterous manipulation~\cite{Guzey2023,van2015learning,lambeta2020digit,funabashi2022multi} and tactile servoing~\cite{tian2019manipulation}.
While these methods allow learning a controller from raw sensor data, they often struggle to generalize to new situations that were not part of the training dataset.
Building on the strengths and limitations of both approaches, we observe that \textbf{Model-based} and \textbf{Model-free} approaches seek to capture manipulator-object dynamics, which depend on intrinsic physical properties (such as mass and friction) and the direction, intensity, and location of interaction forces at the manipulator-object interface.
This suggests that accurate force estimation across the entire manipulator surface would significantly enhance both approaches.
For instance, force estimation is essential both for predicting object dynamics in model-based control and for enforcing safety constraints in learning-based approaches, particularly during manipulation of fragile objects.
Moreover, physical features can serve as structured inputs to learning algorithms~\cite{Koenig2022}, improving data efficiency while maintaining the flexibility of data-driven approaches.
This integration of physical principles and learning-based methods promises to improve both robustness and interpretability in tactile-based manipulation.
Although force estimation is crucial for both precise manipulation and safety guarantees, existing methods either require complex calibration procedures or suffer from slow inference times, limiting their practical application in online control. As a result, most control schemes reduce tactile feedback to stability metrics rather than directly using it to control interaction forces.

In this work, we present a unified approach for calibrating and estimating 3D forces across entire tactile sensor arrays, enabling their direct integration in an interaction force controller. 
Our method provides force reconstruction, from \textit{uncalibrated} measurements of a distributed magnetic tactile sensor system covering the Allegro hand, working seamlessly across different sensor geometries (flat and curved). 
Our contributions are the following:
\begin{itemize}
\item \textbf{(1)} A data-efficient approach for full-array force calibration, demonstrating high accuracy and real-time performance across diverse tactile sensor geometries
\item \textbf{(2)} Integration of tactile-based force estimation in an explicit interaction force control loop, demonstrating its effectiveness for force regulation.
\end{itemize}
\section{BACKGROUND}
\textbf{Estimating forces from tactile sensors poses significant challenges} due to calibration complexities, nonlinear responses, and sensor deformations.
Existing approaches to force calibration can be broadly categorized into element-wise calibration methods for tactile arrays and vision-based approaches for optical tactile sensors. 
Single-element magnetic tactile sensors with force calibration were initially developed~\cite{Tomo2016Design, Paulino2017}, followed by several studies on force calibration for magnetic tactile sensing arrays~\cite{Tomo2016Modular, Tomo2018Covering, Tomo2018New, Sathe2023}. The earliest approach proposed element-by-element force calibration for flat arrays~\cite{Tomo2016Modular}, which was later adapted for the ICub hand~\cite{Tomo2018New} and extended to curved arrays~\cite{Tomo2018Covering, Sathe2023}. 
A different method for simultaneous force calibration of four magnetic sensing elements was proposed for fingertip sensors~\cite{Kristanto2019}, although its calibration setup restricted fingertip movement relative to the ground truth sensor.
These works represent significant breakthroughs in magnetic tactile array development and their integration into robotic systems. 
However, their force calibration approaches present certain limitations. The requirement for sensor detachment makes regular recalibration impractical. Additionally, as noted in~\cite{Tomo2018New}, the time-consuming nature of data collection for shear force measurement restricted testing to a single chip. 
Although high accuracy was achieved (Mean Absolute Error of 0.21, 0.16, and 0.44 [N] for X, Y, and Z axes respectively~\cite{Sathe2023}), two critical aspects remain unaddressed: the robustness of force measurements against different shapes and surface properties due to identical training and testing conditions, and the potential interference from other magnetic tactile arrays present in a fully covered robotic hand.
Our method on the other hand allows for 3D force (normal and shear) calibration of the whole array based on a single data collection step with arrays mounted on a robotic hand. Besides we evaluate our force estimation against different surfaces and geometries.~\cite{Su2015} and~\cite{Navarro2015Active} address force calibration of multi-modal (static pressure and resistive) tactile sensors. 
In~\cite{Navarro2015Active}, while force magnitude can be estimated through its relationship with static pressure, the method cannot determine force direction.~\cite{Su2015} proposes a comprehensive method to estimate force magnitude and direction from a resistive tactile array and draws a benchmark of the performances of different models. 
The estimated forces are then used in a grasp controller but the force feedback is used in conjunction with a slip detection metric as a stability metric, to avoid object slippage and not in an interaction force controller. Also, the type of sensors used in those works cannot be used to fully cover a robot manipulator.
Optical tactile sensors which constitute an important part of the literature often rely on Vision models~\cite{Lin20209DTact,Yuan2017} and face different challenges: they require data-intensive learning, limiting their generalization to new surface types, and often don't specify inference speed requirements for online control.
The challenge of force estimation is further complicated by the inherent properties of optical tactile sensors contact pads. 
Those contact pads are made from very soft or elastic materials, which exhibit non-linearities in the relationship between applied forces and sensor output. 
These behaviors, governed by material properties, contact dynamics, and surface geometry, affect force distribution and cause cross-axial signal coupling.
While some works have attempted to address these challenges through off-robot data collection or detailed finite element models~\cite{Ma2019, Su2015}, these approaches limit online applicability and effectiveness in dynamic tasks.

\textbf{The integration of tactile feedback in interaction control} introduces another set of challenges. 
Most existing approaches use tactile information primarily as a stability metric or compute force heuristics for grasp control, rather than explicitly controlling interaction forces~\cite{Deng2020}. 
While~\cite{Kim2024} demonstrate stable grasping using tactile-based interaction control, its framework lacks quantitative validation of force estimation accuracy, leaving uncertainty about the controller's relationship to physical interaction forces. 
In~\cite{zhang2019effective}, although force estimation is integrated in the control loop, the large sensors are not well-suited for robotic hands, and the estimation is used primarily for stability monitoring rather than for explicit force control. 
Recent work~\cite{Kitouni2024fingertip} proposes a fingertip interaction force direction regulator, but relies on heuristic representations due to lack of proper sensor calibration, controlling interaction-related features rather than actual forces.
While conventional tactile-based control schemes use contact forces primarily for stability monitoring, we integrate our estimation in an interaction control loop that explicitly regulates internal forces of manipulated objects. 
Another critical consideration for interaction control is the required control bandwidth. While typical interaction control loops operate at high frequencies (200Hz~\cite{Amanhoud2020}), many tactile sensing solutions have limited bandwidth. Optical tactile sensors, for instance, typically operate at lower frequencies (25Hz for GelSight~\cite{Yuan2017}, 60Hz for Digit~\cite{lambeta2020digit}). This limitation stems from both the image processing requirements and mechanical constraints of sensor interface~\cite{Kappassov2015}. 
These challenges—limited bandwidth, lack of explicit force control, and unreliable force estimation—highlight the need for an integrated approach that combines accurate force estimation with high-bandwidth sensing capabilities.

\begin{figure}[!h]
    \centering
    \includegraphics[width=0.4\linewidth]{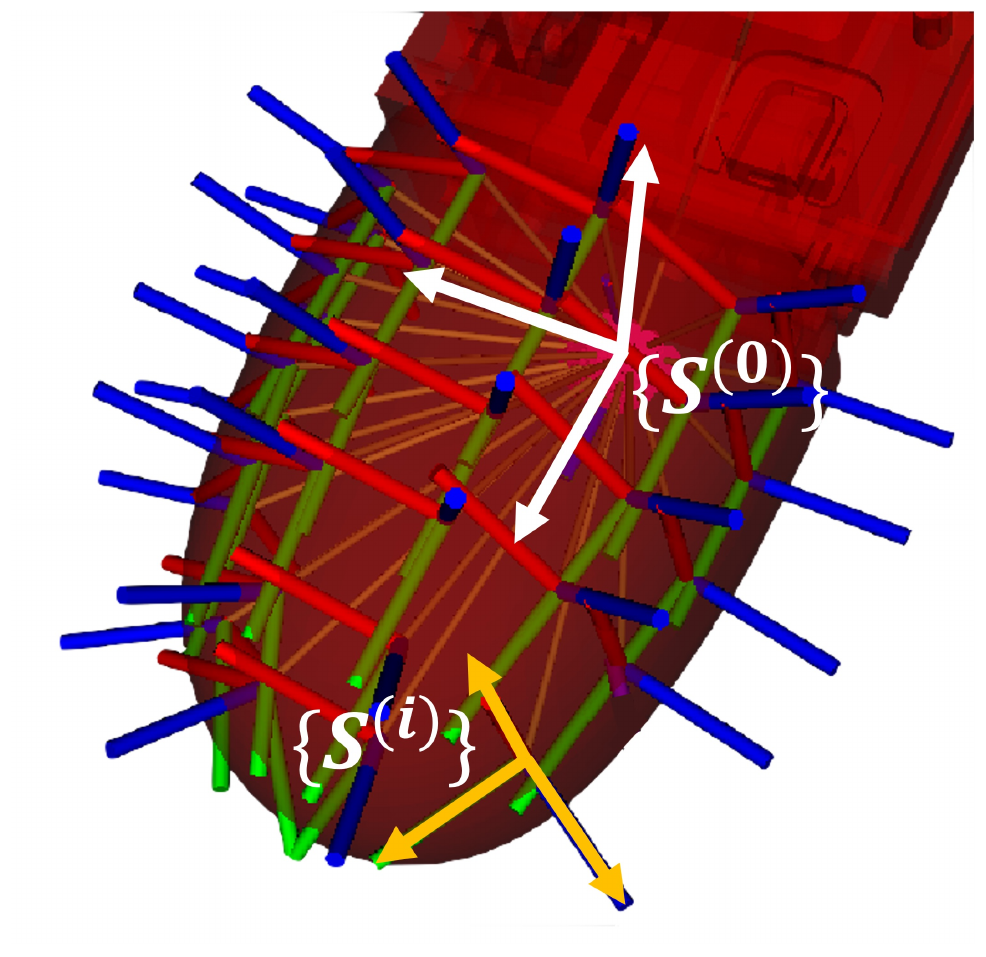}
    \caption{Fingertip with a curved tactile array. Each taxel (30 here) measures its activation in local frame $\{S^{(i)}\}$; $\{S^{(0)}\}$ is the finger tip frame. Each RGB frame represents a taxel.}
    \label{fig:frames}

    \vspace{-0.5cm}
\end{figure}

\section{PROPOSED METHOD}

Let us consider a single sensor array of taxels on a fingertip of a robotic hand (Fig.~\ref{fig:frames}).
In our formulation, each taxel measures \textit{uncalibrated} sensor activities in its local frame.
The measurement for the $i$th taxel is called $\boldsymbol{x}^{(i)} \in \mathds{R}^3$, specified in its local frame $\{S^{(i)}\}$, and we note $\{S^{(0)}\}$ the common frame attached to the last rigid link of the finger. 
Besides, $\boldsymbol{R}^0_{i} \in \text{SO}(3)$ is the -- known -- rotation matrix between $\{S^{(i)}\}$ and $\{S^{(0)}\}$. Using these matrices, we can obtain the sum of all tactile activities in $\{S^{(0)}\}$ ($\boldsymbol{z} \in \mathds{R}^3$):
\begin{equation}
    \boldsymbol{z}  = \sum_{i=1}^n \boldsymbol{R}^0_i \boldsymbol{x}^{(i)}.
    \label{eq:tactile-projection}
\end{equation}
There are $n$ taxels in a sensor array, which we concatenate to build $\boldsymbol{x} \in \mathds{R}^{3n}$, representing the raw measured tactile activities. In this work, we neglect dynamics induced by deformation, and hysteresis. Then, we assume that the tactile activities are the results of the interaction of the sensor array with the environment:
\begin{equation}
    \boldsymbol{x} = \mathcal{G}(\boldsymbol{f}, \tau, \boldsymbol{x}_{c},\boldsymbol{\kappa}, \gamma) + \boldsymbol{x}_{\epsilon},
    \label{eq:tactile-hypothesis}
\end{equation}
with $\mathcal{G}$ a smooth, nonlinear function, $\boldsymbol{f} \in \mathds{R}^3$ the resultant interaction force, $\tau \in \mathds{R}$ the moment around the surface normal at $\boldsymbol{x}_{c} \in \mathds{R}^3$ (the contact location on the sensors' surface), and $\boldsymbol{\kappa} \in \mathds{R}^2$ the principal curvatures of the touched surface, all expressed in $\{S^{(0)}\}$; 
$\gamma \in \mathds{R}$ is the sensor temperature and $\boldsymbol{x_{\epsilon}} \in \mathds{R}^{3n}$ represents unmodeled dynamics. For the ground truth, we rely on a calibrated external sensor,  which provides an unbiased estimation of $\boldsymbol{f}$, noted $\boldsymbol{\tilde{f}} \in \mathds{R}^3$.
In this work, we solve the inverse problem: $\boldsymbol{x}$ is given and we estimate $\boldsymbol{f}$ using a parametric model  $\boldsymbol{\mathcal{H}}: \mathds{R}^{3n} \rightarrow \mathds{R}^3$.
More precisely:
 \begin{equation}
     \boldsymbol{\hat{f}} = \boldsymbol{\mathcal{H}}(\boldsymbol{x};\boldsymbol{\theta}),
     \label{eq:general_parametric_model}
 \end{equation}
with $\boldsymbol{\hat{f}} \in \mathds{R}^3$ the estimated interaction forces using tactile sensor data and $\boldsymbol{\theta} \in \mathds{R}^p$ with $p$ the number of model parameters.
In the following, we present the formulation for five candidate models, suitable for tactile sensor arrays.\\

\textbf{Rotation-assisted linear model (M1)} exploits the prior knowledge about the rotation matrices between common and local frames. The model assumes that there exists a linear relationship with offset between $\boldsymbol{z}$ -- obtained via (\ref{eq:tactile-projection}) -- and the interaction force $\boldsymbol{f}$ in the $\{S^{(0)}\}$ frame.
\begin{equation}
    \boldsymbol{\hat{f}}_{M_1} = \boldsymbol{A}_{M_1}\boldsymbol{z} + \boldsymbol{b}_{M_1},
    \label{eq:m1}
\end{equation}
with $\boldsymbol{A}_{M_1} \in \mathds{R}^{3\times3}$ and $\boldsymbol{b}_{M_1} \in \mathds{R}^3$ the model parameters leading to $p_{M_1}=12$.

\textbf{Rotation-assisted quadratic model (M2)} assumes a quadratic relationship between $\boldsymbol{z}$ and the real interaction force, as in \cite{Tomo2018Covering}. To this end, we define an extended regressor vector $\boldsymbol{z'} \in \mathds{R}^9$, as:
\begin{equation}
    z' = [z_1, z_2, z_3, z_1^2, z_2^2, z_3^2, z_1z_2, z_2z_3, z_1z_3]^T,
\end{equation}
leading to the linear (w.r.t. the parameters) model:
\begin{equation}
    \boldsymbol{\hat{f}}_{M_2} = \boldsymbol{A}_{M_2}\boldsymbol{z'} + \boldsymbol{b}_{M_2},
    \label{eq:m2}
\end{equation}
with $\boldsymbol{A}_{M_2} \in \mathds{R}^{3\times9}$ and $\boldsymbol{b}_{M_1} \in \mathds{R}^3$ resulting in $p_{M_2}=30$. While relying on the prior knowledge about the arrangement of the taxel (${S^{(i)}}$) as in M1, this model has more potential to capture nonlinearities in tactile activities.

\textbf{Raw input linear model (M3)} does not rely on the prior knowledge about the orientation of the taxel and directly takes the raw sensor activities as input: 
\begin{equation}
    \boldsymbol{\hat{f}}_{M_3} = \boldsymbol{A}_{M_3} \boldsymbol{x} + \boldsymbol{b}_{M_3},
    \label{eq:m3}
\end{equation}
with $\boldsymbol{A}_{M_3} \in \mathds{R}^{3\times3n}$ and $\boldsymbol{b_{M_3}} \in \mathds{R}^3$ the model parameters ($p_{(M3)}=9n+3$).
Regardless of the linearity, this model considers a full coupling behavior; i.e., the estimated force in one axis depends on all taxel activities in all three dimensions. Considering such coupling improves the robustness of the estimation w.r.t. sensor deformation.

\textbf{Fully connected multi-layer perceptron (M4)} allows capturing nonlinearities in tactile measurements. We consider a fully connected neural network with one hidden layer. It has an input layer of size  $\dim(\boldsymbol{x}) = 3n$ with \textit{ReLu} activation, a hidden layer of size $16$ with \textit{ReLu} activation, and a linear output layer of size $\dim(\boldsymbol{\hat{f}}) = 3$. This model has $p_{M_4}=48n+69$ parameters.

\textbf{Convolutional neural network (M5)} exploits the sensor arrays' spatial information, via a convolutional neural network \cite{Lecun1998}. Treating the array as an RGB image is a common practice in the literature \cite{Funabashi2020, Meier2016}.
We assume that the $n$ taxels are arranged as a grid of $h$ by $w$ where $h  w = n$.
The network is composed of two convolutional layers (kernel size $3 \times 3$, output channels: 6 and 12) with \textit{ReLu} activation, one 2D batch normalization layer with 12 features, one more convolutional layer with kernel size $2 \times 2$, 24 output channels, and \textit{ReLu} activation, one dense layer (16 neurons) with \textit{ReLu} activation and a linear output layer of size $\dim(\boldsymbol{\hat{f}}) = 3$.
In this model, $p_{(M5)}=2479$.\\

\section{Experimental Setup}

\begin{figure}[h]
    \centering
    \includegraphics[width=1\linewidth]{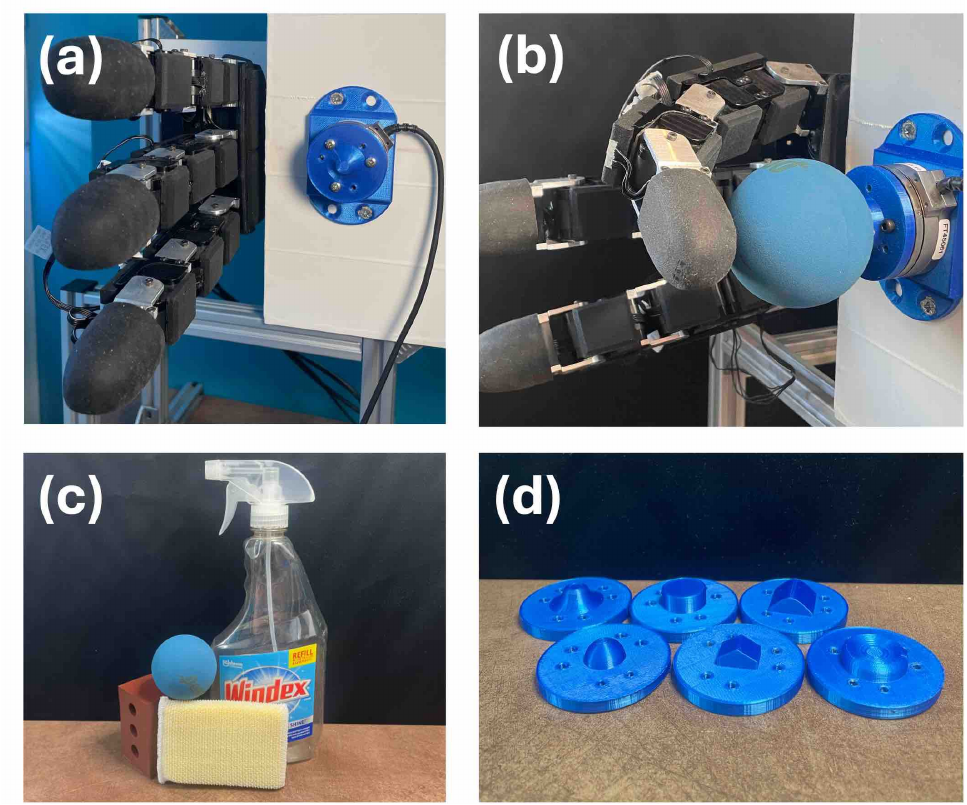}
    \caption{\textbf{(a) Hardware setup:} we use a uSkin covered Allegro hand and an ATI 45mini force torque sensor for calibration. Both are mounted on a rigid chassis. \textbf{(b) Interaction control setup:} Object inserted between fingertip and reference force torque sensor. \textbf{(c) Selected YCB objects:} Foam brick, Softball, Abrasive sponge and Bleach cleanser. \textbf{(d) Push-plates:} Button, Flat, Fillet, Convex, Edge and Spherical.}
    \vspace{-0.4cm}
    \label{fig:exp-setup}
\end{figure}

This section outlines the experimental hardware setup and provides a comprehensive overview of the procedures followed throughout each step of the experiment, from data collection to model validation and closed loop integration. 
First, the data collection process is described, followed by an explanation of the specific data processing and model fitting techniques. 
The evaluation of the models on offline data is presented next, followed by their validation using online data. 
Finally, we discuss the selection of the force controller and closed loop integration of the force estimation.\\

We utilize a 16-degree-of-freedom Allegro hand equipped with Xela Robotics' uSkin \cite{Tomo2016Modular}, including curved fingertip sensor arrays \cite{Tomo2018Covering}. 
The hand features 18 tactile arrays: 4 curved fingertip arrays (30 taxels each), 11 square phalanx arrays (16 taxels each), and 3 rectangular arrays (24 taxels each).
For reference force measurements, we employ an ATI Mini45 force-torque sensor. 
The Allegro hand is mounted on a fixed chassis, with a plate secured perpendicular to the hand. 
The force-torque sensor is attached to the plate at a position that ensures that as many tactile arrays as possible can be pressed against it as can be viewed in Fig.~\ref{fig:exp-setup}. 
Six PLA push-plates with different shaped indenters are 3D printed. Five of them are used for data collection and one is reserved for online validation. 
The force-torque sensor measures forces in frame $\{F\}$ with the origin at the center of the indenter's surface, ensuring minimal torque application during data collection. This point is consistent across the six plates.
The transformation between the Allegro Hand's base frame $\{B\}$ and $\{F\}$ is measured once and is then supposed static. Four objects from the YCB~\cite{Calli2015} dataset with different stiffnesses are selected (Fig.\ref{fig:exp-setup} (c)) and reserved for the closed loop control experiments.\\
\vspace{-5pt}
\subsection{Data collection and force estimation}

Prior to each data collection session, the hand and tactile sensors are powered on for a minimum of 30 minutes, to allow the system to reach thermal equilibrium, thus minimizing the impact of temperature on the tactile signal. 
Data synchronization, frame transformations, and time alignment are handled via the ROS Noetic framework \cite{ros2009}. 
The time varying rotation matrix $\boldsymbol{R}^{(0)}_F(t)$ between the array's common frame $\{S^{(0)}\}$ and the force-torque sensor frame  $\{F\}$ is computed during the length of acquisition from the robot's URDF file and joint positions. 
A human operator presses the array against the extrusion of the push plate, varying both the applied forces' magnitude and directions.\\

For each plate geometries, except the spherical one which is kept for online validation, four sequences of 100 seconds each are collected. Each dataset is composed of tactile data, measured forces, and frame transformations, recorded at 100 Hz, imposed by the tactile sensor's maximum sampling frequency. Signals are processed as follows: \textbf{(1)} Low-pass filtered with a phase-compensated and 10 Hz cut-off frequency. \textbf{(2)} Resampled to synchronize the time vectors. \textbf{(3)} Force data projected onto the tactile sensor's frame $\{S^{(0)}\}$ using $\boldsymbol{R}^{(0)}_F(t)$. \textbf{(4)} Offsets computed from static segments and subtracted. \textbf{(5)} Standardized using their respective standard deviations. Data collection is repeated for each fingertip and intermediate phalanx array, resulting in 200,000 data points per array.\\

Models M1 to M5 are trained using different optimization algorithms. The linear and quadratic models (M1 to M3) are fitted using the least squares method. A second version of M3, noted M3$\lambda$, is trained with the dampened least squares method, with a damping of $\lambda = 33$, which was found to give the best results. M4 and M5 are trained using PyTorch’s \cite{Pytorch2019} implementation of the Adam optimizer \cite{Adam2014}, with a mean squared error (MSE) loss function. For these models, we used a batch size of 256, a learning rate of $2.5 \times 10^{-4}$, and trained them for 40 epochs. Various neural network architectures were tested, and the best-performing ones were selected for this work.\\
 
Each model is trained iteratively on data from 4 of the 5 plates and tested on the remaining plate, in a 5-fold cross-validation fashion.
Offline performance on each unseen plate is then averaged an reported in the result section. 
While this provides useful insight on how well models generalize to new data, it does not fully reflect their effectiveness in real-world applications. 
We therefore conduct two additional phases of validation, both involving real-robot experiments. The first phase involves testing the models online, where the sensor array is pressed against the unseen spherical plate (Fig\ref{fig:exp-setup}(b)) mounted on the external force-torque sensor, and errors between $\boldsymbol{\tilde{f}}$ and $\boldsymbol{\hat{f}}$ are computed and recorded. The second phase is a closed-loop validation, where the models are integrated in a controller for the Allegro Hand. Here, the model's output directly affects future inputs, enabling us to assess its performance in dynamic, real-time manipulation tasks.
\vspace{-5pt}
\subsection{Closed loop control}

\begin{figure}[!t]
    \centering
    \includegraphics[width=1\linewidth]{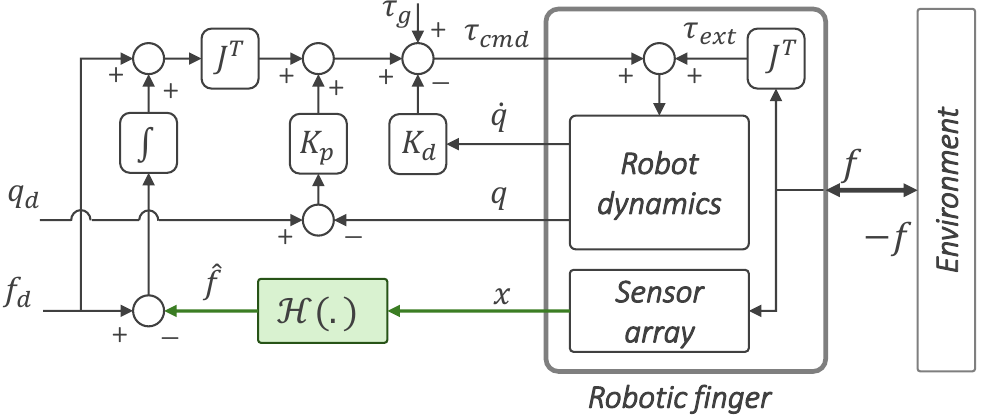}
    \caption{Closed loop task space force controller. The tactile-equipped hand is in rigid contact with the environment; $\boldsymbol{x}$ is the tactile observation and $\boldsymbol{\mathcal{H}}$ the function s. that $\boldsymbol{\mathcal{H}}(\boldsymbol{x}) = \boldsymbol{\hat{f}}$.}
    \vspace{-0.4cm}
    \label{fig:controller_block_diagram}
\end{figure}

Our proposed closed-loop interaction force controller is shown in Fig.~\ref{fig:controller_block_diagram}, where the main control objective is to regulate the interaction force; i.e., minimize the error between the desired force $\boldsymbol{f}_d$ and the estimated force $\boldsymbol{\hat{f}}$. The controller is applied to one finger with 4 D.O.F, $\boldsymbol{q}\in\mathds{R}^4$. Once more, the controller is tested on the spherical plate, which is not part of the training set and the four selected deformable YCB objects shown in Fig.~\ref{fig:exp-setup}(b). The applied force is measured by the FT sensor and is used for error computation and evaluation.

The control law is:
\begin{equation}
    \boldsymbol{\tau}_{cmd} = \boldsymbol{J}^T(\boldsymbol{f}_d - \boldsymbol{K}_i\int \boldsymbol{e}_f dt) + \boldsymbol{\tau}_g - \boldsymbol{K}_d\dot{\boldsymbol{q}} - {\boldsymbol{K}_p}(\boldsymbol{q} - \boldsymbol{q}_{d}),
    \label{eq:full-controller}
\end{equation}
with $\boldsymbol{\tau}_{cmd} \in \mathds{R}^4$ the joint torque command, $\boldsymbol{J} \in \mathds{R}^{3\times 4}$ the translation Jacobian of the end effector, $\boldsymbol{\tau}_g \in \mathds{R}^4$ the gravity compensation torque and $\boldsymbol{K}_p$ and $\boldsymbol{K}_d$ the $4\times4$ diagonal matrix for joint position error.

We base our control law on the method proposed in \cite{Lee2017}, which is implemented on an Allegro hand without force or tactile feedback. 
Our force feedback is incorporated as an integral correction term.
We only use an integral and no proportional nor derivative term for the force feedback, since it has been shown\cite{Wilfinger1994} that for a robot in contact with a rigid surface, this approach is more robust. By setting $\boldsymbol{K_i} = \boldsymbol{0}$, we obtain an open-loop variant of the controller.
Additionally, proportional and derivative joint position control terms are summed and $\boldsymbol{q}_{d}$ is selected so that the controlled fingertip remains in contact with the push-plate or object.
 
To prevent the integral term from becoming excessively large, an integral error rate limiter and an anti-windup mechanism are implemented. Additionally, \(\boldsymbol{\tau}_{cmd}\) is saturated at 0.35 \([ \text{Nm} ]\), since we observed that exceeding this value causes the actuators to overheat and enter a temperature throttling mode, which reduces their power output. When any component of \(\boldsymbol{\tau}_{cmd}\) is saturated, the remaining components are adjusted proportionally to ensure that the direction of the resulting applied force remains unchanged. Both the control loop and the estimation run at 100 Hz.

We designed five experimental scenarios to evaluate force controller performance, each repeated five times per controller variant. Specific models for the closed-loop controllers were selected based on performance metrics described in Section~\ref{sec:results}. In experiment \textbf{(1)}, the finger pushed against the spherical push-plate directly mounted on the force sensor. We test open-loop and closed-loop vairants of the controller. For experiments \textbf{(2)}--\textbf{(5)}, we placed different YCB objects between the finger and the force sensor, with each object taped to the sensor. Although the estimated finger forces and measured sensor forces were not collocated in these cases, we ensured that the contact points (finger-to-object and object-to-sensor) lay on parallel tangential planes. For these object-interaction experiments, only the closed-loop controllers were tested.

\section{Results} \label{sec:results}

This section presents the performance of our models for force estimation using tactile observations. We first evaluate all models in an offline benchmark, then assess their real-time performance on live data. Finally, we use the estimations for force feedback in the closed-loop controller (Eq.~\ref{eq:full-controller}) and compare performances with an open-loop version of the controller (Eq.~\ref{eq:full-controller} with $\boldsymbol{K_i} = \boldsymbol{0}$). The code, dataset and additional video demonstrations are accessible here. [\url{https://tinyurl.com/tac2f}].
\vspace{-5pt}
\subsection{Offline model benchmark}
Table~\ref{tab:offline-results} summarizes performance of our force estimation models on curved fingertip arrays and flat phalanx arrays, using relative error $e_r$ and mean absolute error $\hat{e}$:
    \noindent\begin{minipage}{0.24\textwidth}
\begin{equation}
e_r = \frac{1}{3}\sum_{i=1}^{3}|\frac{(\tilde{f}_{i} - \hat{f}_i)}{\tilde{f}_{i}}|\label{eq:rel-error}
\end{equation} 
    \end{minipage}%
    \begin{minipage}{0.1\textwidth}\centering
    \end{minipage}%
    \begin{minipage}{0.25\textwidth}
\begin{equation}
\hat{e} = \frac{1}{3}\sum_{i=1}^{3}|(\tilde{f}_{i} - \hat{f}_i)|\label{eq:abs-error}
\end{equation}
    \end{minipage}\vskip0.32cm
Data points with $\tilde{f_i} < 0.5 [\text{N}]$ are excluded from Eq.~\ref{eq:rel-error} to avoid singularities. 
Each model is trained on data from 4 push-plate geometries and tested on the 5th. $e_r$ and $\hat{e}$'s mean and standard deviation are computed. This process is repeated 5 times, with each of the five push-plate geometries serving as the test set exactly once (5-fold cross-validation). $e_r$ and $\hat{e}$'s mean and standard deviation are finally averaged across the 5 folds and reported in Tab.~\ref{tab:offline-results}.\\
\vspace{-5pt}
\begin{table}[!h]
    \centering
\begin{tabular}{@{}lccc@{}}
\toprule
\textbf{Model} & \begin{tabular}[c]{@{}c@{}} \textbf{Avg $e_r$ [\%]}\\ \end{tabular} & \begin{tabular}[c]{@{}c@{}} \textbf{Avg $\hat{e}$ [N]}\\ \end{tabular} & \begin{tabular}[c]{@{}c@{}} \textbf{Mean $\tau$ [ms]}\\ \textbf{Flat}\end{tabular} \\    
\toprule
\multicolumn{4}{c}{\textbf{Curved Array}} \\
\midrule
M1          & $52\pm21.04$ & $0.75\pm0.34$ & $0.393$\\
M2          & $54\pm23.57$ & $0.72\pm0.35$ & $0.691$\\
M3          & $34\pm19.47$ & $0.36\pm0.23$ & $0.047$\\
M3$\lambda$ & $33\pm19.14$ & $0.35\pm0.23$ & $0.044$\\
M4          & $24\pm14.80$ & $0.27\pm0.18$ & $0.051$\\
M5          & $23\pm12.04$ & $0.25\pm0.15$ & $0.370$\\
\midrule
\multicolumn{4}{c}{\textbf{Flat Array}} \\
\midrule
M1          & $39\pm20.13$ & $0.37\pm0.20$ & $0.053$\\
M2          & $38\pm19.12$ & $0.34\pm0.20$ & $0.082$\\
M3          & $38\pm20.84$ & $0.31\pm0.20$ & $0.046$\\
M3$\lambda$ & $37\pm20.67$ & $0.31\pm0.19$ & $0.040$\\
M4          & $27\pm15.79$ & $0.23\pm0.15$ & $0.050$\\
M5          & $27\pm14.36$ & $0.22\pm0.14$ & $0.230$\\
\bottomrule                                                       
\end{tabular}
\caption{Model benchmark for one fingertip curved array and one phalanx flat array. Details in Section III. M3$\lambda$: Linear model with dampening factor. $\tau$ is the computation time for a single prediction.}
\label{tab:offline-results}
\end{table}

The results demonstrate that models not assuming fixed taxel orientation (M3 - M5) outperform those that incorporate this information (M1, M2) on the curved fingertip array. This performance advantage is not due to the absence of prior knowledge but rather to the models' ability to account for sensor deformation and nonlinearities. By avoiding rigid assumptions about geometry, these models are better equipped to capture the complex interactions between taxels under deformation. 
In contrast, on the flat phalanx array, where $\{S^{(i)}\}$ all have the same orientation, the main difference between M1 - M2 and M3 - M5 lies in the treatment of taxel data: M1 and M2 sum taxel activities, while M3 - M5 process each taxel independently. In both cases, M4 and M5 which are based on neural network perform best, due to their flexibility which allows the models to capture the fact that different stimuli due to indenter geometry variation can result in the same force measurement. 

Based on these findings, we proceed with further evaluation using models M4 and M3$\lambda$. M3$\lambda$ is chosen for its simplicity and robust performance across both curved and flat arrays. 
M4 is also chosen because of its higher performance across both arrays. Both models have the advantage of fast inference time (0.04 [ms] for M3$\lambda$ and (0.05 [ms] for M4), making them suitable for running 18 parallel instances for full-hand surface force estimation.\\
\vspace{-5pt}
\subsection{Online evaluation}
Models M4 and M3$\lambda$ were tested online for 200 seconds on the curved index fingertip array on the spherical push-plate (excluded from training data). Both models were first retrained on the complete dataset, having previously used only four of five splits.
During the test, the index fingertip is pressed against the push-plate, mounted on force sensor, and the corresponding data are recorded. $e_r$ and $\hat{e}$ are computed and reported in Tab.~\ref{tab:online-results}.
Results show that M4 outperformed M3$\lambda$ in terms of both relative and absolute error metrics, demonstrating strong robustness in dynamic, real-world scenarios. M3$\lambda$ still achieves exploitable results and is kept for closed-loop evaluation.\\

\begin{table}[h]
\centering
\begin{tabular}{ @{}lccc@{}}
\toprule

\textbf{Model} & \begin{tabular}[c]{@{}c@{}} \textbf{Average $e_r$}\\ \textbf{[\%]} \end{tabular} & \begin{tabular}[c]{@{}c@{}} \textbf{Avg $\hat{e}$}\\ \textbf{error [N]} \end{tabular} \\ 

\toprule
M3$\lambda$  & $25.70\pm11.00$ & $0.31\pm0.14$ \\ 
M4  & $20.90\pm9.80$ & $0.21\pm0.10$ \\ 
\bottomrule
\end{tabular}
\caption{Results for real time force estimation with M4 and M3$\lambda$ on the index fingertip sensor.}
\label{tab:online-results}
\end{table}
\vspace{-15pt}
\subsection{Closed loop evaluation}

\begin{figure*}[!t]
    \centering
    \begin{subfigure}[b]{0.49\linewidth}
        \centering
        \includegraphics[width=\linewidth]{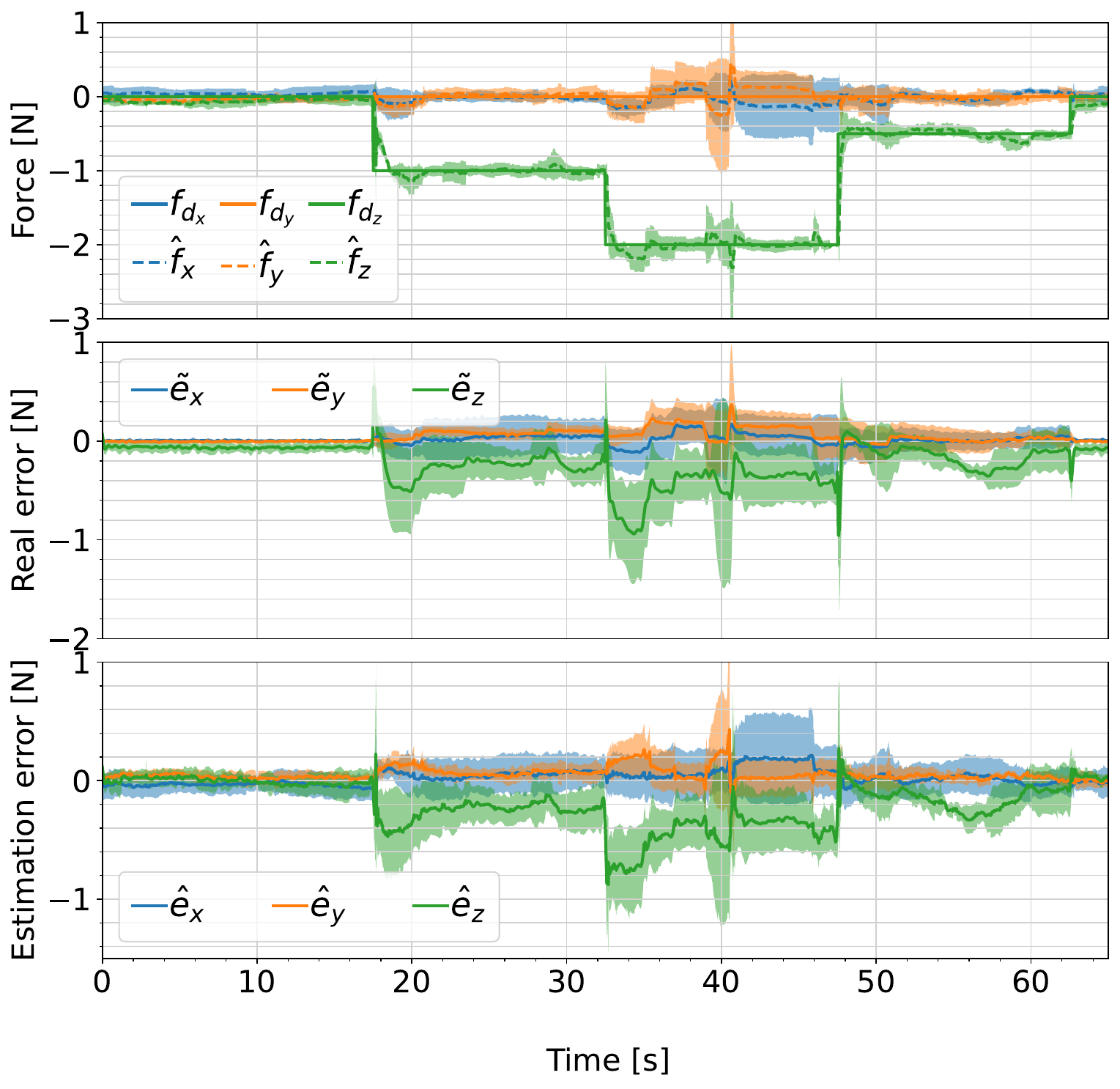}
        \caption{}
        \label{fig:closed-loop}
    \end{subfigure}
    \hfill
    \begin{subfigure}[b]{0.49\linewidth}
        \centering
        \includegraphics[width=\linewidth]{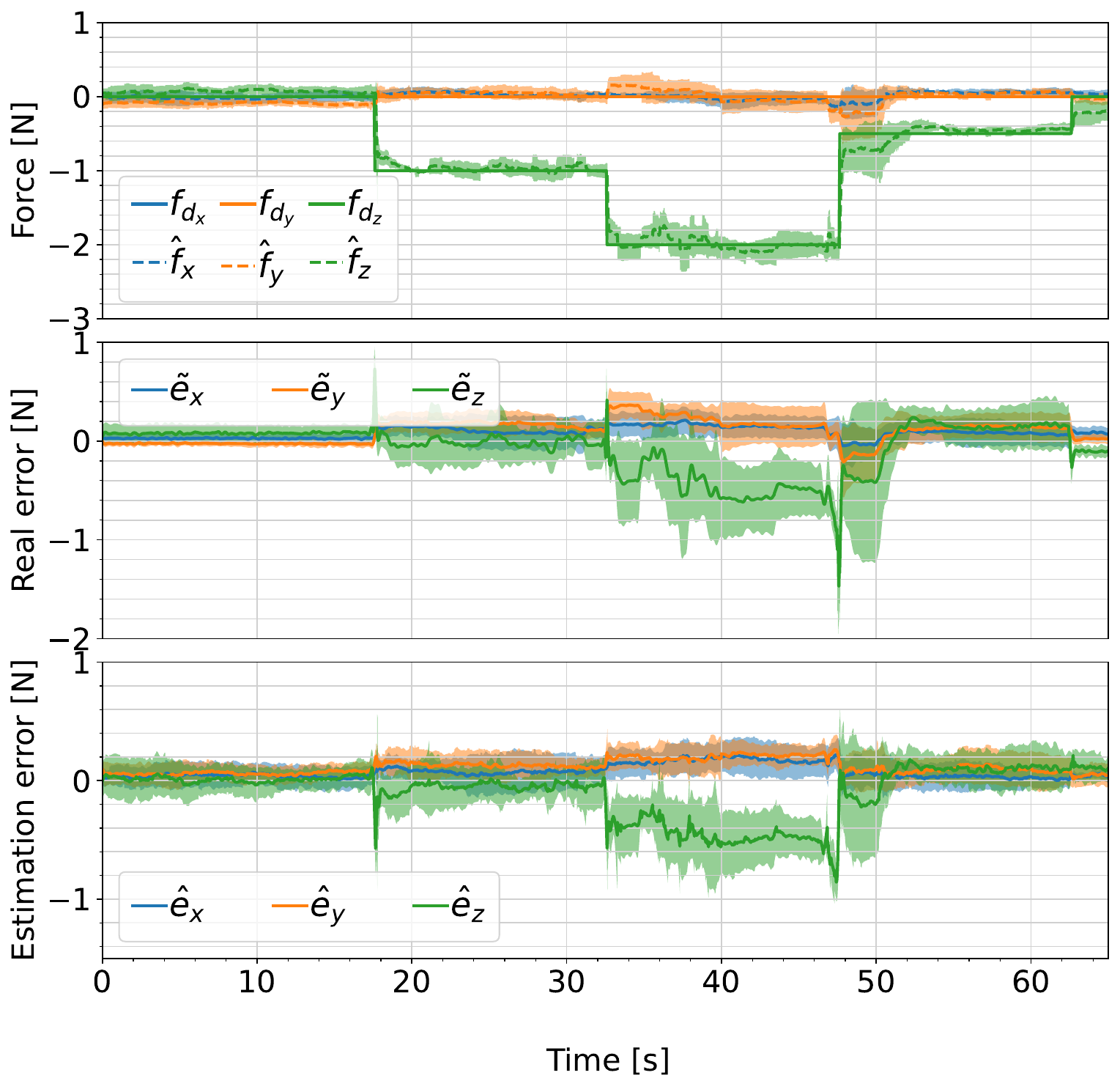}
        \caption{}
        \label{fig:soft-obj}
    \end{subfigure}
    \caption{\textbf{(Left)} Closed loop force controller performances. The force estimation is computed with $M3\lambda$. \textbf{(Right)} Closed loop force controller performances, with a soft object in between the finger and the force sensor (YCB foam brick). The force estimation is computed with $M3\lambda$. Each experiment is repeated 5 times, the solid line represents the average and the aura represents the standard deviation}
    \label{fig:controller}
\end{figure*}

We integrated models M4 and M3$\lambda$ into the controller presented in Eq.~\ref{eq:full-controller} and compared their performance against the open-loop variant (Eq.~\ref{eq:full-controller} with $\boldsymbol{K}_i = \boldsymbol{0}$). Results are reported in Tab.\ref{tab:control-results}. The controller was applied to the index finger, with the fingertip as the end effector, and all forces specified in the $\{F\}$ frame. We tested all three controller versions with the same 70[s] force command sequence $\boldsymbol{f_d}$ to evaluate both force tracking performance and estimation quality in static scenarios.
For closed-loop controllers, we calculated relative error $e_r$ (Eq.~\ref{eq:rel-error}), absolute error $\hat{e}$ (Eq.~\ref{eq:abs-error}), estimate tracking error $e_{track} = \frac{1}{3}\sum_{i=1}^{3}|(f_{d_i} - \hat{f}_i)|$ (error between the desired force and force estimate), and true tracking error $\tilde{e} = \frac{1}{3}\sum_{i=1}^{3}|(f_{d_i} - \tilde{f}_i)|$ (error between desired force and ground truth measurement). For the open-loop controller, we only report $\tilde{e}$ since other error metrics depend on force estimation $\boldsymbol{\hat{f}}$.
All error metrics were averaged over a time window from t=16[s] to t=70[s], excluding the initial phase ($\boldsymbol{f_d} = \boldsymbol{0}$) to avoid skewing the results.

\definecolor{darkgreen}{RGB}{0,180,0}

\begin{table}[!t]
\centering
\setlength{\tabcolsep}{5pt}
\begin{tabular}{ @{}lcccc@{}}
\toprule
\textbf{Model}  & 
\begin{tabular}[c]{@{}c@{}} \textbf{Avg $e_{r}$}\\ \textbf{error [\%]} \end{tabular} &
\begin{tabular}[c]{@{}c@{}} \textbf{Avg $\hat{e}$}\\ \textbf{error [N]} \end{tabular} &
\begin{tabular}[c]{@{}c@{}} \textbf{Avg $e_{track}$}\\ \textbf{error [N]} \end{tabular} &
\begin{tabular}[c]{@{}c@{}} \textbf{Avg $\tilde{e}$}\\ \textbf{error [N]} \end{tabular} \\
\toprule
(1)-O.L.&$--$&$--$&$--$&$\textcolor{red}{0.46\pm0.15}$\\
(1)-M3$\lambda$&$20.32\pm7.08$&$0.12\pm0.08$&$0.06\pm0.05$&$0.12\pm0.08$\\
(1)-M4&$4.86\pm6.86$&$0.04\pm0.03$&$0.05\pm0.05$&$\textcolor{darkgreen}{0.06\pm0.06}$\\
(2)-M3$\lambda$&$\textcolor{darkgreen}{12.36\pm8.02}$&$0.14\pm0.09$&$0.06\pm0.04$&$0.16\pm0.08$\\
(2)-M4&$12.54\pm7.89$&$0.14\pm0.09$&$0.07\pm0.04$&$0.14\pm0.09$ \\
(3)-M3$\lambda$&$\textcolor{darkgreen}{14.25\pm6.60}$&$0.13\pm0.07$&$0.07\pm0.05$&$0.13\pm0.08$\\
(3)-M4&$23.87\pm7.16$&$0.25\pm0.16$&$0.12\pm0.05$&$0.22\pm0.14$\\
(4)-M3$\lambda$&$\textcolor{darkgreen}{16.04\pm7.97}$&$0.18\pm0.11$&$0.05\pm0.05$&$0.17\pm0.13$\\
(4)-M4&$26.93\pm22.01$&$0.25\pm0.29$&$0.13\pm0.10$&$0.25\pm0.25$\\
(5)-M3$\lambda$&$\textcolor{darkgreen}{12.42\pm7.08}$&$0.13\pm0.07$&$0.07\pm0.05$&$0.13\pm0.07$\\
(5)-M4&$19.60\pm5.55$&$0.17\pm0.10$&$0.06\pm0.03$&$0.15\pm0.12$\\
\bottomrule
\end{tabular}
\caption{Controller results summary. (1) Spherical Indenter; (2) Foam brick; (3) Softball; (4) Abrasive sponge; (5) Bleach cleanser. For (1) lowest $\tilde{e}$ are highlighted in \textcolor{darkgreen}{green} and highest in \textcolor{red}{red}. For the remaining, lowest $e_r$ are highlighted in \textcolor{darkgreen}{green}.}
\label{tab:control-results}
\end{table}

As shown in Tab.~\ref{tab:control-results}, both M4 and M3$\lambda$ models significantly reduced $\tilde{e}$ compared to the open-loop controller. When testing on the spherical push-plate (experiment \textbf{(1)}), the M4 model exhibited superior performance, achieving the lowest estimation and tracking errors. However, across experiments \textbf{(2)}-\textbf{(5)} with YCB objects, the M3$\lambda$ model consistently outperformed M4, with the exception of the foam brick test (2) where both models showed comparable results.

This performance pattern suggests that while M4 demonstrates higher accuracy on objects similar to those in the training distribution, M3$\lambda$ possesses superior generalization capabilities when faced with out-of-distribution objects. The closed-loop system with M3$\lambda$ achieved an average $\tilde{e}$ reduction of approximately 74\% compared to the open-loop approach and an $\hat{e} = 0.12\pm0.08 [N]$, highlighting its potential for real-time force control across diverse object interactions.

Further analysis reveals that our estimator performs particularly well on signals with slow dynamics, as evidenced by the superior results of M4 in Tab.~\ref{tab:control-results} (1) compared to Tab.~\ref{tab:online-results}. This demonstrates the model's capacity to maintain accuracy in steady-state force application scenarios, where precise force control is crucial. 

The robustness of the M3$\lambda$ model in handling varying contact areas and surface curvatures suggests it is more resilient to object property variations outside the training distribution. This adaptability is critical for real-world applications, such as soft object grasping and manipulation, making M3$\lambda$ particularly suitable for diverse task-oriented scenarios where interaction with previously unseen objects is required. Its consistent performance across both rigid and deformable surfaces validates its versatility for tasks involving complex interactions with different materials and geometries.

\section{Discussion}

This study introduces a novel method for on-hand calibration and force estimation using magnetic tri-axial tactile sensors, with a particular focus on its application to closed-loop force control. Our approach removes the need for sensor pre-calibration, significantly simplifying integration with robotic manipulators and enabling real-time applications. The effectiveness of our method is validated not only through comparison of tactile force estimates with reference measurements but also through its successful integration into a force control system.
The closed-loop force controller utilizing our M3$\lambda$ model demonstrated a 74\% reduction in tracking error compared to the open-loop approach and low average true tracking error ($\tilde{e}$), ranging from $0.12$ to $0.17$ [N]. Notably, while the M4 model exhibited superior performance on objects similar to the training distribution, M3$\lambda$ showed high generalization capabilities across various out-of-distribution objects, highlighting its robustness for real-world applications. This adaptability is critical for tasks involving interactions with diverse materials and geometries, as confirmed by its consistent performance across both rigid and deformable surfaces.
Beyond force control, our approach has broader implications for applications like teleoperation and soft robotics, where real-time tactile feedback is essential. In teleoperation, the enhanced force feedback precision could improve control and task execution, while in soft robotics, the ability to reliably estimate forces on deformable surfaces enhances adaptability during object manipulation. A major strength of our method is its adaptability to various hardware platforms and tactile sensor types.
Some limitations persist, particularly in dynamic conditions. Furthermore, in interaction control experiments (2)--(5), the non-collocated reference force measurements and estimates likely degraded our average estimation performance, as estimated forces were directly compared to reference measurements potentially perturbed by soft object internal dynamics during transients.
An important insight from this study is that neural network performance for this type of task depends more strongly on training set composition than linear models do. Our real-time calibration method adapts to sensor deformation, providing more reliable force estimation while eliminating labor-intensive pre-calibration procedures. This approach demonstrates that accurate force control can be achieved through adaptive estimation techniques, paving the way for scalable, robust solutions for real-world robotic manipulation tasks.\\

\section{Acknowledgements}
We would like to thank Prof. Guillaume Morel for his help in the implementation of the control loop.



\bibliographystyle{IEEEtran}
\begin{small}
\def\BIBdecl{\setlength{\itemsep}{0pt}}
\def\url#1{}
\def\urlprefix{} 
\def\urlsuffix{} 
\bibliography{IEEEabrv,references}
\end{small}

\end{document}